\def\year{2020}\relax
\documentclass[letterpaper]{article} % DO NOT CHANGE THIS
\usepackage{aaai20}  % DO NOT CHANGE THIS
\usepackage{times}  % DO NOT CHANGE THIS
\usepackage{helvet} % DO NOT CHANGE THIS
\usepackage{courier}  % DO NOT CHANGE THIS
\usepackage[hyphens]{url}  % DO NOT CHANGE THIS
\usepackage{graphicx} % DO NOT CHANGE THIS
\usepackage{graphicx}  % DO NOT CHANGE THIS
\usepackage{amsfonts}
\usepackage{amsmath}
\usepackage{multirow}
\title{CIAN: Cross-Image Affinity Net for Weakly Supervised Semantic Segmentation}
\author{Junsong Fan,\textsuperscript{\rm 1,2,4} Zhaoxiang Zhang,\textsuperscript{\rm 1,2,3,4}\thanks{Corresponding author} Tieniu Tan,\textsuperscript{\rm 1,2,3,4}\\
\Large \textbf{Chunfeng Song,\textsuperscript{\rm 1,2,4} Jun Xiao\textsuperscript{\rm 4}\footnotemark[1]}\\ % All authors must be in the same font size and format. Use \Large and \textbf to achieve this result when breaking a line
\textsuperscript{\rm 1}Center for Research on Intelligent Perception and Computing, CASIA\\ %If you have multiple authors and multiple affiliations
\textsuperscript{\rm 2}National Laboratory of Pattern Recognition, CASIA\\
\textsuperscript{\rm 3}Center for Excellence in Brain Science and Intelligence Technology, CAS\\
\textsuperscript{\rm 4}University of Chinese Academy of Sciences\\
\{fanjunsong2016, zhaoxiang.zhang\}@ia.ac.cn, \{tnt, chunfeng.song\}@nlpr.ia.ac.cn, xiaojun@ucas.ac.cn
}
\begin{document}

\maketitle

\begin{abstract}
Weakly supervised semantic segmentation with only image-level labels saves large human effort to annotate pixel-level labels. Cutting-edge approaches rely on various innovative constraints and heuristic rules to generate the masks for every single image. Although great progress has been achieved by these methods, they treat each image independently and do not take account of the relationships across different images. In this paper, however, we argue that the cross-image relationship is vital for weakly supervised segmentation. Because it connects related regions across images, where supplementary representations can be propagated to obtain more consistent and integral regions. To leverage this information, we propose an end-to-end cross-image affinity module, which exploits pixel-level cross-image relationships with only image-level labels. By means of this, our approach achieves 64.3\% and 65.3\% mIoU on Pascal VOC 2012 validation and test set respectively, which is a new state-of-the-art result by only using image-level labels for weakly supervised semantic segmentation, demonstrating the superiority of our approach. 
\end{abstract}

% %%%%% SECTION1:introduction
\section{Introduction}

Semantic segmentation provides per pixel predictions for a given image. Recently, fully convolutional network (FCN) based methods \cite{fcn,deeplab-v2,deeplab-v3} have achieved impressive performance. However, these deep methods need large scale datasets with precise pixel-level annotations for training \cite{pascal-voc,coco}, which is quite expensive to obtain.
To alleviate the difficulty of collecting data for training, weakly supervised learning (WSL) \cite{wsl-review} is proposed for semantic segmentation. It makes use of weak annotations for training, e.g. bounding boxes \cite{boxsup,wsl-ss-bbox}, sparse scribbles \cite{scribblesup,scribble-rw}, and image-level class labels \cite{sec,erasing,multi-dilation,affinity-net,seed-growing}. In this paper, we focus on the most challenging problem by only using image-level labels for semantic segmentation.

\begin{figure}[t]
\begin{center}
\includegraphics[width=0.8\linewidth]{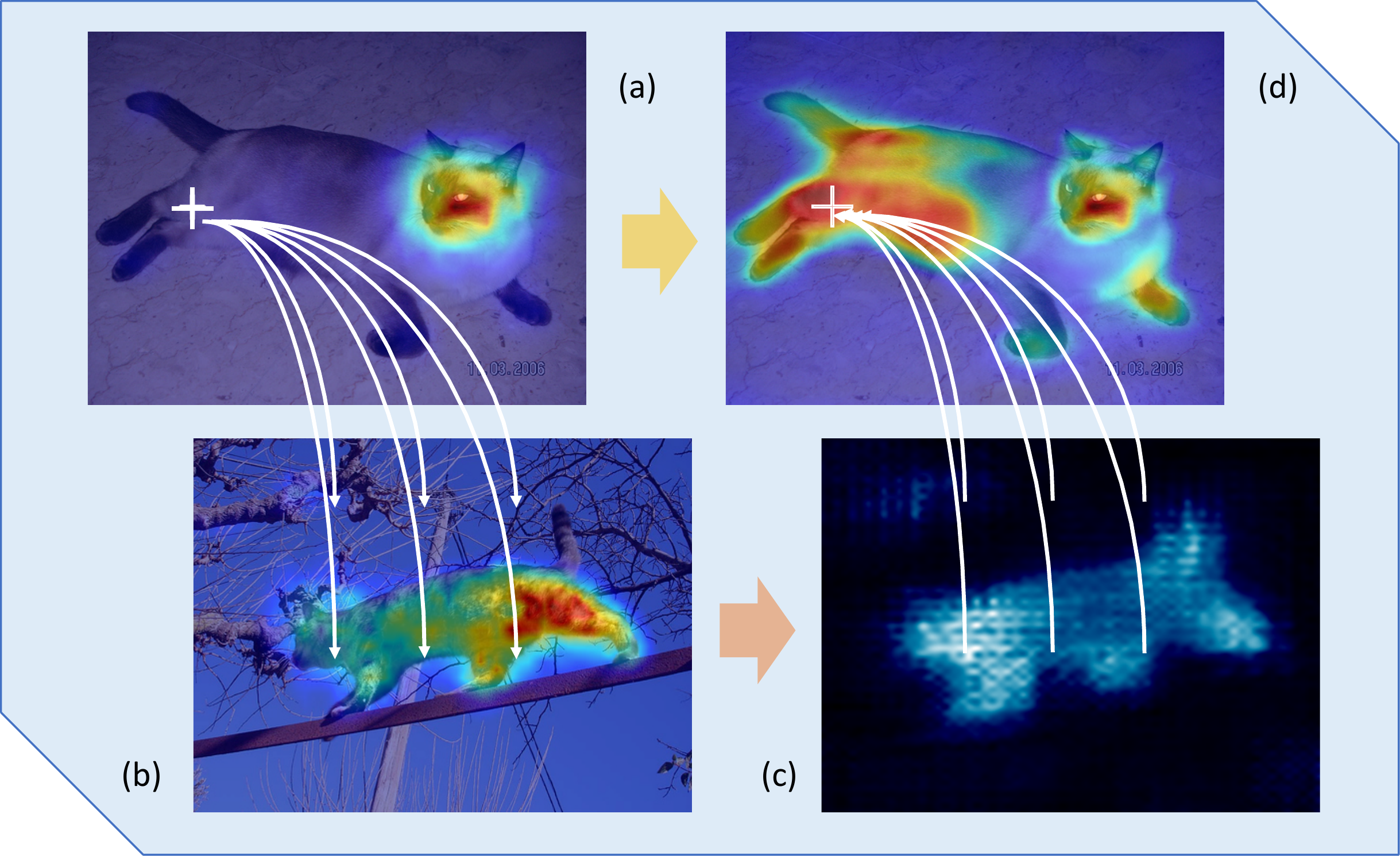}	
\end{center}
\caption{Illustration of the cross-image affinity. (a) the raw activation of a given image. (b) the activation of a reference image of the same class. (c) the affinity map from the marked query pixel in a) to reference b). (d) the activation after retrieving supplementary information according to the affinity.}
\label{fig:motivation}
\end{figure}

The main difficulty of weakly supervised semantic segmentation is to recover the precise spatial information from only image tags. To this end, existing works usually rely on attention mechanisms, e.g. CAM \cite{cam}.
However, these attention maps are derived from classification networks, which focus on the classification precision instead of the targets' integrity, thus only the most discriminative regions are obtained, which are sparse and incomplete.

To tackle this problem, Wei et al. \cite{erasing} adopt an iterative erasing strategy to mine complementary seeds. In the following works \cite{multi-dilation}, they also apply multi-dilation convolution blocks to expand the seeds. Ahn and Kwak \cite{affinity-net} train additional pixel-level affinity net to complete the seeds. Huang et al. \cite{seed-growing} dynamically fill in the undefined regions by seed region growing algorithm. Kolesnikov et al. \cite{sec} and Briq et al. \cite{simplex} take advantage of additional constraints to regularize the predictions. A common characteristic of these methods is that the images are treated independently of each other.

Contrastingly, we propose that the cross-image relationship is also vital for mining complete regions for weakly supervised segmentation. To intuitively understand this concept, see Fig. \ref{fig:motivation}, there are two images of the same class, and only partial regions are activated for each of them. We leverage the pixel-wise affinities and retrieve complementary information from one image to another, then more integral regions can be obtained. More formally, there are three major benefits of introducing the cross-image relationships:

Firstly, the cross-image relationship helps to provide supplementary information for identifying the pixels. Based on the supplementary components, related features can be refined and amplified to address some of the ambiguity and/or false predictions. 
Secondly, this explicit relationship helps the network to learn more consistent representations across the whole dataset. This is because the affinity module propagates related representations across different images, and a consistent one will be approached finally.
Thirdly, through the cross-image relationship, the labels can be directly shared across a group of images, making better use of the valuable weak supervision. 

Based on the motivation of building cross-image relationships, we propose an end-to-end cross-image affinity module, which can be directly plugged into existing segmentation networks. We name it cross-image affinity net (CIAN). CIAN explicitly models the pixel-level relationships among different images, efficiently leverages the relationships to refine the original representations and obtains more integral regions for segmentation.
We conduct thorough experiments to demonstrate the effectiveness of the proposed approach. Our approach achieves new state-of-the-art results of weakly supervised semantic segmentation by only using image-level labels, with 64.3\% mIoU on Pascal VOC 2012 \cite{pascal-voc} validation set, and 65.3\% on the test set. In summary, the main contributions are as follows:

\begin{itemize}
    \item We firstly propose to explicitly model the cross-image relationship for weakly supervised semantic segmentation. An end-to-end cross-image affinity module is proposed to provide supplementary information from related images. By means of this, more integral regions can be obtained for weakly supervised segmentation.
    \item Extensive experiments demonstrate the usefulness of modeling  cross-image relationships. Besides, we show that our approach is orthogonal to the quality of the seeds, which continually improves the training with even better seeds. Thus it can be potentially combined with future works that generate better seeds to further boost the performance.
    \item With the naive seeds generated by CAM, our CIAN achieves 65.3\% mIoU on the VOC 2012 test set, which is a new state-of-the-art result by only using image-level labels for semantic segmentation, demonstrating the superiority of the approach.
\end{itemize}

% %%%%% SECTION2:related-work

\begin{figure*}[t]
\begin{center}
\includegraphics[width=0.95\linewidth]{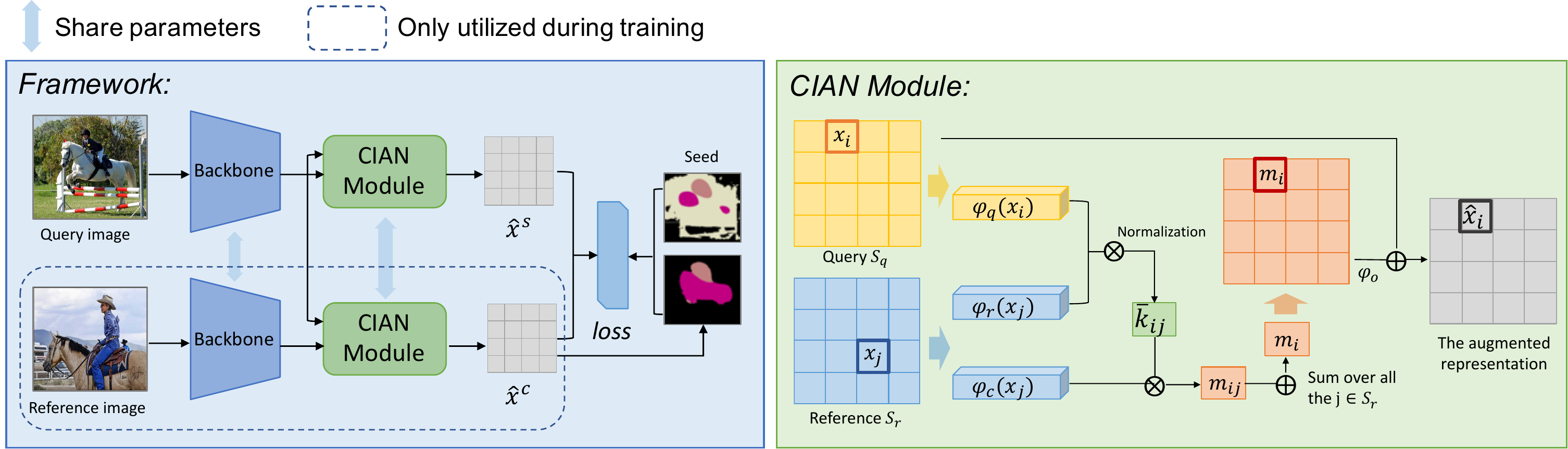}	
\caption{
\textbf{Left}: the framework of CIAN. For simplicity, only one pair is drawn. Embedded features are obtained by a siamese backbone for both query and reference images, following with the cross-image affinity module to augment the features. For testing, reference images are unavailable, thus the query image only pairs to itself.
\textbf{Right}: the proposed cross-image affinity module (CIAN module). The affinity is derived from the query and the reference features, then the query retrieves supplementary information from the reference accordingly.}
\label{fig:framework}
\end{center}
\end{figure*}

\section{Related Work}

\noindent\textbf{Weakly Supervised Semantic Segmentation.}
In this paper, we focus on the image-level label based weakly supervised semantic segmentation. State-of-the-art approaches follow a pipeline of first generating pseudo-masks (seeds) then training segmentation networks. The CAM \cite{cam} is widely adopted as the cornerstone for generating seeds. However, CAM only activates the most discriminative regions, which is incomplete for segmentation.

To alleviate this problem, Wei et al. \cite{erasing} propose to adopt an iterative erasing strategy. MDC \cite{multi-dilation} proposes to merge multiple CAMs with different dilation rates. DSRG \cite{seed-growing} proposes to dynamically fill in the sparse seeds by region growing. Wang et al. \cite{iterative-mine} propose to alternately train a superpixel based classification network and the segmentation network. Other works \cite{simplex,sec} propose some heuristic constraints. The concurrent work FickleNet \cite{ficklenet} randomly drops connections in each sliding window. Although these methods are effective, they ignore the rich relationships across different images, while we prove that the cross-image relationship is effective for obtaining consistent and integral regions for weakly supervised segmentation.

\noindent\textbf{Co-segmentation.}
Co-segmentation aims to predict the common objects' masks for a given group of images \cite{deep-co-seg,deep-obj-coseg,deep-co-saliency,unsup-co-attention,unsup-co-saliency}. This task is related to ours since we also operate on a group of images to learn and utilize the cross-image relationship. The main difference between co-segmentation and weakly supervised segmentation is that co-segmentation focuses on finding class-agnostic masks for universal objects. When testing, it takes as input a group of images, so that common object can be defined. While our weakly supervised segmentation infers single images for known classes. Besides, many co-segmentation approaches are trained by strong pixel-level masks \cite{deep-co-seg,deep-co-saliency}. There is also weakly supervised segmentation by adopting co-segmentation to generate seeds \cite{weak-seg-by-co-seg}, but it is different from ours that our affinity module is an end-to-end component for segmentation networks instead of the seeds.

\noindent\textbf{Pixel-Level Affinity.}
Recent work AffinityNet \cite{affinity-net} samples sparse points by CAM seeds and trains an additional affinity net by metric learning. It is different from ours that our affinity module is an end-to-end component for dynamically sharing information across different images. In form, our method is closely related to the Non-local approaches \cite{non-local,dual-attention,ocnet}, since pixel-level affinity is incorporated. However, Non-local networks focus on the long-range intra-image contexts to discover the hidden structures of a single image, while our approach aims at highlighting common objects and share complementary information across different images to combat the weak label problem.

% %%%%% SECTION3:our-approach

\section{Our Approach}
In this section, we elaborate on all the components of the proposed approach. Following common practice, \cite{sec,multi-dilation,seed-growing,seenet}, we firstly generate initial seeds from image labels, then use them to train the segmentation network which is equipped with our proposed CIAN module. The framework is illustrated in Fig. \ref{fig:framework}.

\subsection{Initial Seeds}
\label{sec:initial-seeds}
Following the common practice \cite{sec,multi-dilation,seed-growing,seenet}, we adopt on-the-shelf CAM \cite{cam} approach to generate initial seeds. It trains a classification network, which has a global-pooling layer right before the last classification layer. After training, it removes the global-average pooling layer and directly applies the classification layer to every pixel in the feature map to obtain the score map. However, because the classification tasks only focus on the most discriminate regions, only sparse and incomplete regions are highlighted. Our proposed CIAN can effectively alleviate the influence of the incomplete seeds with the help of cross-image relationships.

\subsection{The Cross-Image Affinity Module}
\label{sec:affinityModule}
The affinity module models pixel-wise relationships between two images upon high-level representations. For simplicity, we use the term \textit{pixels} to refer spatial-wise vectors of the feature map. 
Let $S_q$ and $S_r$ denote the sets of pixel indices of query image $q$ and reference $r$, respectively. For any pair of pixels $\{(x_i, x_j) | i\in S_q, j\in S_r\}$ from the two images, the affinity score $k_{ij}$ is modeled by:

\begin{equation}
k_{ij} = \exp\left(\varphi_q(x_i)\cdot\varphi_r(x_j)\right)
\label{eq:dot-product}
\end{equation}
where $\varphi_q$ and $\varphi_r$ are learnable functions implemented by neural network layers, which can be seen as generalized kernel functions to enhance the encoding flexibility. 

The next step is to retrieve supplementary information from reference $x_j$ to query $x_i$. To this end, we first compress useful information by function $\varphi_c$, and weight it by the corresponding affinity score to get the \textit{message} $m_{ij}$:

\begin{equation}
m_{ij} = \bar{k}_{ij}\varphi_c(x_j)	
\label{eq:weight-info}
\end{equation}
where, $\bar{k}_{ij}$ is the normalized version of affinity $k_{ij}$, to ensure that the sum of all the weights of the reference pixels is a unit, $\bar{k}_{ij}=k_{ij}/\sum_{j\in S_r}k_{ij}$.

Finally, all the messages from different reference pixels to the query $x_i$ are summed together, normalized by function $\varphi_o$, and merged into the original representation $x_i$:

\begin{equation}
m_i = \sum_{j\in S_r} m_{ij}
\label{eq:sum-message}
\end{equation}
\begin{equation}
\hat{x}_i = x_i + \varphi_o(m_i)
\label{eq:merge-message}
\end{equation}

The above $\hat{x}_i$ is the so-called cross-affinity augmented representation. This process is repeated for all the pixels in the query image. The final classification layer takes as input augmented representations and outputs the segmentation results. The whole process is illustrated in Fig. \ref{fig:framework}.

On the one hand, meaningful affinity is forced to be learned to fit the existing seed supervision. On the other hand, the learned affinity bridges different images together to provide supplementary and/or complementary information. By means of this, the augmented representation and the affinity prompt each other and can be learned simultaneously.

\subsection{Multiple Pairs}
\label{sec:multiple-pairs}
The above affinity module operates on a pair of two images.
It can be easily extended to formulate relationships among multiple reference images and a single query image.

Given the query image $q$ and its $N$ reference partners $\{r^{(h)} | h=1, ..., N\}$, we compute all the messages from the reference images to $q$ according to Eq. \ref{eq:sum-message}, which are denoted as $\{m_i^{(h)}| i\in S_q; h=1, ..., N\}$.  Then, before adding them into the corresponding raw representation $x_i$, we merge all the messages from the multiple pairs. For example, it can be implemented by maximum function:

\begin{equation}
m_i = \max_{h\in \{1, ..., N\}} m_i^{(h)}
\label{eq:merge-pair}
\end{equation}

Finally, $m_i$ is normalized by $\varphi_o$ and added to the original $x_i$, as in Eq. \ref{eq:merge-message}. Other merging functions, e.g. average function, are also available.

\subsection{Training Loss}
\label{sec:training-loss}

\paragraph{Cross-Entropy Loss.}
Following the common practice of semantic segmentation \cite{deeplab-v1,deeplab-v2,fcn}, we adopt the pixel-wise cross-entropy loss to train the segmentation network. Use $f_c$ to denote the final softmax classification layer of the segmentation network, the class probability of pixel $\hat{x}_i$ is then obtained by $f_c(\hat{x}_i) \in \mathbb{R}^C$, where $C$ is the number of classes. For image $q: \hat{x}=\{\hat{x}_i|i\in S_q\}$, the cross entropy loss is defined as:

\begin{equation}
	L_{ce}(\hat{x}) = -\frac{1}{|S'_q|} \sum_{i\in S'_q} y^T_i \log f_c(\hat{x}_i)
\label{eq:loss-ce}
\end{equation}
where $y_i$ is the onehot pseudo-label for pixel $x_i$ obtained from the initial seeds. We use $S'_q$ to denote the set of valid pixels for training in image $q$, since there are may pixels not assigned with labels by the seeds. The unassigned pixels are just ignored during training.

\paragraph{Completion Loss.}
As discussed above, the cross-affinity module augmented representations can provide more complete object estimations. Thus, it can be utilized to online generate pseudo-masks to compensate the sparsity of the initial seeds. To this end, we generate the online pseudo-label $\hat{y}_i\in\mathbb{R}^{C}$ from the prediction $f_c({\hat{x}_i})$:

\begin{equation}
\hat{y}_{i,l}=\mathbb{I}\left[\left(\arg\max f_c(\hat{x}_i)=l\right) \mathbf{AND} \left(l\in L_q\right)\right]
\label{eq:online-generate}
\end{equation}
where, $L_q$ is the set of image-level labels for image $q$, $\mathbb{I}[\cdot]$ is the index function and equals 1 if the statement is true. The meaning of Eq. \ref{eq:online-generate} is that the $l$-th element of $\hat{y}_i$ is 1 iff it matches the prediction $f_c(\hat{x}_i)$ and the image-level labels.

Finally, we use $\hat{y}$ to optimize all the pixels in the query image $q$, and the whole procedure is named completion loss:

\begin{equation}
	L_{cp}(\hat{x}) = -\frac{1}{|S''_q|} \sum_{i\in S''_q} \hat{y}^T_i \log f_c(\hat{x}_i)
\label{eq:loss-cp}
\end{equation}
where, $S''_q$ is the set of effective pixels according to $\hat{y}$.

\paragraph{The Overall Loss.}
To ensure there do exist usable information across images, we need common-class pairs for training. However, it is hard to sample pairs with identically the same classes, since there can be multiple classes in a single image thus faces a combinatorial explosion problem. Therefore, we relax to sample images with at least one common class as pairs. To reduce the influence of possibly unmatched classes, we utilize the \textit{self-affinity}, which simply adopts the above affinity module to the image itself, i.e. $x_q=x_r$. We denote the representations augmented by the cross-affinity and the self-affinity as $\hat{x}^{c}$ and $\hat{x}^{s}$, respectively. The overall loss is computed as:
\begin{equation}
    L = L_{ce}(\hat{x}^c) + L_{ce}(\hat{x}^s) + L_{cp}(\hat{x}^c) + L_{cp}(\hat{x}^s)
\label{eq:loss-all}
\end{equation}

Another important reason to use the self-affinity is that during testing we cannot make pairs since the deployed model should be able to address single images. Directly removing the affinity residual incurs heavy distribution change of representations, thus is unfeasible. Instead, we address this problem by augmenting the representation with the self-affinity during testing. To this end, $L_{ce}(\hat{x}^s)$ and $L_{cp}(\hat{x}^s)$ are minimized to further reduce the training and testing gap.

% %%%%% SECTION4:experiments

\section{Experiments and Analysis}
\label{sec:experiments}
\subsection{Datasets}
We evaluate our proposed method on Pascal VOC 2012 segmentation benchmark \cite{pascal-voc}. This is the standard dataset for weakly supervised semantic segmentation. It has 20 foreground classes and one background class. A single image may contain multiple classes. Following the common practice \cite{erasing,multi-dilation,seed-growing}, we use the expanded set collected by Hariharan et al. \cite{voc-extra-segmentation}, i.e., there are 10582 training images, 1449 validation images, and 1456 testing images. In our experiments, only the image class labels are used for training. The performance is evaluated by mean intersection over union (mIoU) of all the 21 classes.

\subsection{Implementation Details}
\paragraph{Initial Seeds.}
As aforementioned, we adopt CAM to generate initial seeds. Specifically, it uses ImageNet pre-trained VGG16. To obtain larger maps, we replace the last two pooling layers with stride 1 and use dilation rate 2 in the Conv5 block. We train the CAM with the multi-class sigmoid loss with learning rate $1e^{-3}$. We normalize the CAM into range $[0, 1]$ and generate foreground regions by threshold 0.3. Following related works \cite{sec,multi-dilation,seed-growing,seenet}, we use an off-the-shelf saliency model \cite{salient} to generate background seeds by threshold 0.06. Finally, all the remaining unassigned pixels and conflictual assignments are abandoned and marked as ignored.

\paragraph{CIAN Module.}
We choose Deeplab-V2-Largefov \cite{deeplab-v2} framework for segmentation.
The CIAN module takes as input the feature maps right before the classification layer.
$\varphi_q$, $\varphi_r$ and $\varphi_c$ are all implemented by single $1\times 1$ convolution layers.
To speedup the computation, $\varphi_q$ and $\varphi_r$ halve the feature dimensions, $\varphi_r(x)$ and $\varphi_c(x)$ are spatially downsampled by max-pooling with stride 2.
During training, we randomly sample reference images for each query image.
We experimentally find that a single reference image for each query is adequate for learning cross-image relationships, more pairs bring negligible improvement.
This may because the pairs are randomly sampled and all the potential combinations can be visited along the training process.
To stabilize the training, the image to itself is also adopted as a pair and merged by Eq. \ref{eq:merge-pair}.
Since the affinity is unreliable during the initial training stage, a zero-initialized batch normalization layer is attached to $\varphi_o$ before adding it to original representations.
Following \cite{seed-growing}, we also adopt the retraining strategy, i.e. generate predictions by currently trained network, and take these predictions as pseudo-labels to train the network with another round.

\begin{table}
\centering
\resizebox{0.8\columnwidth}{!}{
\begin{tabular}{l|c|c}
	\hline
	 & \# Params  & \# FLOPs \\
	\hline
	Vanilla Deeplab & 42.4 M & 72.5 G \\
	Baseline		& 50.8 M & 88.2 G \\
	Ours			& 50.8 M & 89.7 G \\
	\hline
\end{tabular}}
\caption{Comparison of the number of parameters and computation complexity. The values are based on ResNet101.}
\label{table:complexity}
\end{table}

\begin{figure}[t]
\begin{center}
\includegraphics[width=0.95\linewidth]{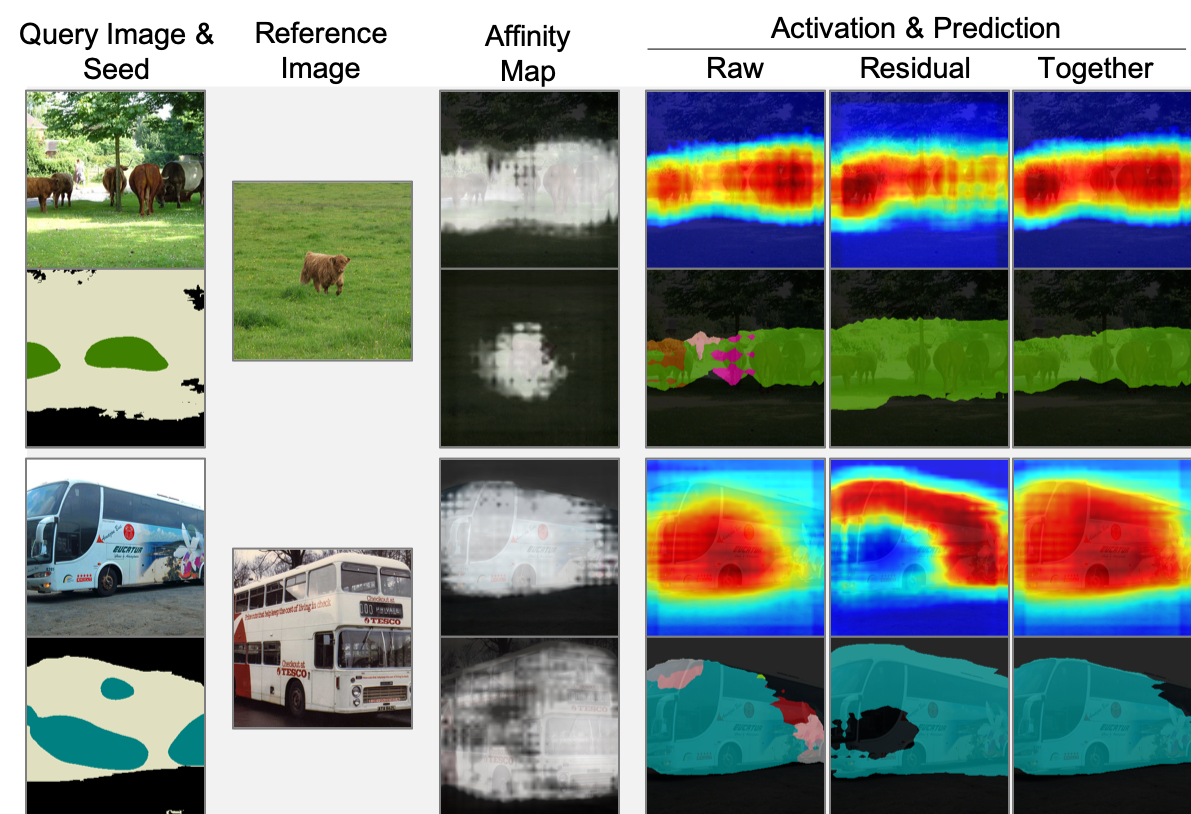}
\end{center}
\caption{Visualization of the affinity. Two typical pairs are illustrated. The first two columns are the query and reference respectively. The third column is the average affinity map for the self- and the cross-affinity respectively. The last three columns show the activation and predictions of the raw, residual, and augmented representations, respectively.
}
\label{fig:visualization}
\end{figure}

\paragraph{Reproducibility.}
All the backbones are pre-trained by ImageNet, and the newly added layers are initialized by random Normal with the standard deviation 0.01.
We adopt the SGD optimizer with an initial learning rate $5e^{-4}$ and momentum 0.9, which is poly-decayed by power 0.9.
We use batch size 16 to train 20 epochs with randomly cropped images of size 321.
Standard data augmentation, i.e., random cropping, scaling, and horizontal flipping are adopted.
CRF \cite{crf} with default parameters for post-processing. Codes are implemented with MXNet \cite{mxnet}, and are available at: https://github.com/js-fan/CIAN.

\subsection{Computation Complexity}
Our CIAN module works at the top layer with relatively small spatial size, thus the overhead is marginal compared with the standard segmentation networks, as shown in Table \ref{table:complexity}.
Besides, the computation complexity of our CIAN is the same as the self-affinity augmented baseline during testing, and the overhead only happens during training.

\subsection{An Intuitive View of The Cross-Image Affinity}
To help the reader further understand how the cross-image affinity helps, we visualize the learned maps in Fig. \ref{fig:visualization}.
Since the affinity component is an addable residual to the raw representation, we visualize the activation of the raw representation, the cross-affinity residual, and the final augmented representation, respectively, as shown in the last three columns in Fig. \ref{fig:visualization}.

We can see that the affinity maps (in the third column) focus on correlated target object regions.
The first example shows that the spatial distribution of the retrieved residual's activation is similar to the raw. After merging, the representation is further strengthened, thus some of the false predictions can be avoided.
The second example shows that the cross-image residual retrieves complementary activation, and thus makes more integral final predictions. This case illustrates that different images may be activated with different parts, and the learned cross-image affinity can help to share complementary information among images to achieve more complete and consistent estimations.

\begin{table*}[t]
\begin{center}
\resizebox{\textwidth}{!}{
\begin{tabular}{c| ccccc ccccc ccccc ccccc c |c}
	\hline
	Method & \rotatebox{90}{bkg} & \rotatebox{90}{plane} & \rotatebox{90}{bike} & \rotatebox{90}{bird} & \rotatebox{90}{boat} & \rotatebox{90}{bottle} & \rotatebox{90}{bus} & \rotatebox{90}{car} & \rotatebox{90}{cat} & \rotatebox{90}{chair} & \rotatebox{90}{cow} & \rotatebox{90}{table} & \rotatebox{90}{dog} & \rotatebox{90}{horse} & \rotatebox{90}{motor} & \rotatebox{90}{person} & \rotatebox{90}{plant} & \rotatebox{90}{sheep} & \rotatebox{90}{sofa} & \rotatebox{90}{train} & \rotatebox{90}{tv} & mIoU \\
	\hline
	Baseline 		& 85.0 & 73.7 & 33.5 & 74.6 & 49.1 				& 63.0 & 77.5 & 68.6 & 65.9 & 22.7 &
			   		  63.0 & \textbf{40.2} & 58.4 & 64.5 & 69.9 	& 57.1 & 35.9 & 74.3 & \textbf{34.5} & 61.1 & 50.9 & 58.3 \\
			   		  
	+ CE			& 84.9 & 75.6 & \textbf{34.1} & 72.4 & 45.1		& 65.8 & 79.1 & 67.6 & 68.7 & 21.1 &
					  66.9 & 39.7 & 62.4 & 72.5 & 71.4				& 58.7 & 34.4 & 75.8 & 32.5 & 61.2 & 52.0 & 59.1 \\
	
	+ CP			& 87.1 & 76.0 & \textbf{34.1} & 73.1 & 49.7		& 69.5 & 83.5 & \textbf{73.9} & 77.5 & 27.2 &
					  70.9 & 37.1 & 72.4 & 74.6 & 73.6				& 62.5 & 42.7 & 76.6 & 34.2 & \textbf{63.6} & \textbf{53.5} & 62.5 \\
					  
	+ RT			& \textbf{88.2} & \textbf{79.5} & 32.6 & \textbf{75.7} & \textbf{56.8} 		& \textbf{72.1} & \textbf{85.3} & 72.9 & \textbf{81.7} & \textbf{27.6} &
					  \textbf{73.3} & 39.8 & \textbf{76.4} & \textbf{77.0} & \textbf{74.9}		& \textbf{66.8} & \textbf{46.6} & \textbf{81.0 }& 29.1 & 60.4 & 53.3 & \textbf{64.3} \\
	\hline
\end{tabular}}
\end{center}
\caption{Comparison on the VOC 2012 val set for ablation study. The baseline is only augmented by self-affinity. (+ CE) denotes to adopt cross-image affinity with the cross-entropy loss. (+ CP) denotes to further adopt the completion loss. (+ RT) denotes to adopt the retraining strategy.}
\label{table:ablation-study}
\end{table*}

\begin{table}[t]
\begin{center}
\begin{tabular}{c|cc}
	\hline
	Method & Common-class & Random-class \\
	\hline
	+ CE & \textbf{59.1} & 58.5 \\
	+ CP & \textbf{62.5} & 59.7 \\
	\hline
\end{tabular}
\end{center}
\caption{Comparison on the VOC 2012 val set between the random-class pairs and our common-class pairs. In random-class pairs, images may not have any common classes, while in our common-class pairs they have at least one common class.}
\label{table:random-pairs}
\end{table}

\begin{table}
\begin{center}
\resizebox{\columnwidth}{!}{
\begin{tabular}{c|c| cc|c| cc|c}
\hline\hline
\multicolumn{2}{c|}{\multirow{2}{*}{Ratio}} & \multicolumn{3}{c|}{ResNet101} & \multicolumn{3}{c}{ResNet50} \\
\cline{3-8}
\multicolumn{2}{c|}{} & baseline & ours & Delta 		& baseline & ours & Delta\\
\hline

\multirow{3}{*}{\rotatebox{90}{CRF}}  &0    & 58.3 & 62.5 & + 4.2 		& 57.5 & 60.4 & + 2.9 \\
										 &0.05 & 61.4 & 65.8 & + 4.4 		& 60.2 & 63.5 & + 3.3 \\
										 &0.10 & 62.2 & 67.3 & + 5.1 		& 61.6 & 65.7 & + 4.1 \\
\hline

\multirow{3}{*}{\rotatebox{90}{no CRF}} &0    & 53.8 & 58.1 & + 4.3 		& 52.7 & 56.0 & + 3.3 \\
										 &0.05 & 57.7 & 61.2 & + 3.5 		& 56.4 & 59.4 & + 3.0 \\
										 &0.10 & 60.3 & 64.1 & + 3.8 		& 58.4 & 62.1 & + 3.7 \\
\hline\hline

\end{tabular}}
\end{center}
\caption{Comparison on the VOC 2012 val set with different ratios of strong supervision. The larger ratio corresponds to better seeds. Our approach continually improves over the baseline with different seeds. Results both with or without CRF post-processing are given.}
\label{table:semi-supervised}
\end{table}

\subsection{Ablation Studies}
We conduct thorough experiments to demonstrate the advantage of the CIAN module. By default, ResNet101 is adopted as the backbone, and results are evaluated on Pascal VOC 2012 validation set.

\paragraph{The Effect of The CIAN Module.}
We first prove that the learned cross-image relationship benefits the weakly supervised segmentation task. For fair comparisons, the \textit{baseline} model is also augmented by the affinity module, with the limitation that no cross-image pairs are available, i.e., only self-affinity is applied for the baseline. By means of this, the capacity of the baseline model is the same as our CIAN, and we can conclude that the improvement actually comes from the cross-image relationships exploited during training.

The ablation results are shown in Table \ref{table:ablation-study}. Starting with the baseline, with the naive cross-entropy loss for the cross-image representations (+CE), the model achieves 0.8\% improvement. By further adopting the completion loss with the cross-image representations (+CP), the model achieves another 3.4\% improvement. This improvement is much more than the former because the cross-affinity augmented representations provide more complete pseudo-masks than the initial seeds. Even though the cross-affinity can directly refine the representations, with only sparse regions from the initial seeds, much useful information will be abandoned and does not participate in the optimization due to the enormous number of ignored pixels. Thus, the completion loss is necessary for mining cross-image relationships and fully utilizing the retrieved complementary information. Finally, the retraining strategy recovers boundaries by the CRF refinement when generating predictions, thus the result is further improved by 1.8\% (+RT) and achieves the cutting-edge performance, as shown in Table \ref{table:state-of-the-art}.

\paragraph{The Query-Reference Pairs}
In our CIAN, the query-reference pairs are sampled that there is at least one common class. If not, e.g. assume that we randomly sample reference images to the query without any class label constraint, reliable relationships would not be learned. From this point of view, we can also demonstrate the usefulness of valid cross-image relationships.

To this end, in contrast to the proposed common-class sampling strategy, we train the CIAN model with random-class pairs. The results are summarized in Table \ref{table:random-pairs}. It shows that with random-class pairs, the performance is in line with the baseline model, which is much worse than our common-class counterpart.

We notice that with random-class pairs and completion loss, there is still 1.4\% improvement compared with the baseline. This is because although there are no valid cross-image pairs, the network's online predictions still provide better pseudo-masks than initial sparse seeds. However, without valid cross-image relationships, the completion is limited and inferior. As a comparison, our CIAN with reliable common-class pairs outperforms the baseline by 4.2\%, which is much better than the random-class pair's 1.4\%. This result also reveals that only adopting the online completions (as in Eq.\ref{eq:loss-cp}) without available cross-image relationships does not fully utilize the information.

\paragraph{Orthogonal to The Seed Quality}
We prove that our approach does not rely on occasionally generated initial seeds. Indeed, the CIAN module consistently improves over the baseline with even stronger seeds. Therefore, our approach is orthogonal to those state-of-the-art approaches \cite{multi-dilation,erasing,seenet}, which generate better initial seeds.
To quantitatively assess the orthogonality, we imitate a group of better seeds by randomly substituting a portion of the seeds with the ground truth, which is similar to the setting of semi-supervised learning.
Specifically, 5\% or 10\% of the 10582 training seeds are substituted, respectively.

As shown in Table \ref{table:semi-supervised}, with 5\% and 10\% of the seeds substituted, our CIAN outperforms the baseline by 4.4\% and 5.1\% respectively. Similar improvements are achieved with the ResNet 50 backbone. Our approach consistently brings significant improvement with better seeds. Therefore, this approach can be potentially fused with those works generating stronger seeds.

\subsection{Comparison with State-of-The-Art}
It should be careful to make comparison with other approaches, because they may leverage additional supervision, different pre-trainings, or adopt different backbones. We summarize the state-of-the-art results and list their difference in Table \ref{table:state-of-the-art}.

Our approach with ResNet101 achieves mIoU score of 64.3\% and 65.3\% on VOC12 val and test set respectively, outperforming all of the previous results by only using image-level labels. It is surprising that our result outperforms some early fully supervised works like FCN \cite{fcn}. In spite of the different backbones, our result is comparable with some works with stronger supervisions, e.g. box-supervised \cite{boxsup} and scribble-supervised \cite{scribblesup} approaches.
With the same backbone and training set, our approach outperforms AISI \cite{salient-instance}, which relies on a well-trained instance saliency network. Note that instance saliency is trained by pixel-level annotated instance masks, which is quite costly to obtain.
Besides, our result with ResNet50 is comparable with the state-of-the-arts with ResNet101 \cite{iterative-mine,seed-growing,seenet}, demonstrating the advantage of leveraging cross-image relationships.

\begin{table}[t]
\resizebox{\columnwidth}{!}{
\begin{tabular}{llcc}
	\hline
	Methods & Sup. & val & test \\
	
	\hline\hline
	\multicolumn{4}{l}{\textit{Fully supervised}} \\
	
	FCN\textdagger			{\scriptsize\cite{fcn}}					& {\scriptsize\textit{F.}}				& -	   & 62.2 \\
	Deeplab\textdagger		{\scriptsize\cite{deeplab-v1}}			& {\scriptsize\textit{F.}}				    & 67.6 & 70.3 \\
	
	\hline
	\multicolumn{4}{l}{\textit{Weakly supervised}} \\
	
	BoxSup\textdagger 		{\scriptsize\cite{boxsup}}				& {\scriptsize\textit{L.}+\textit{B.}}	& 62.0 & 64.6 \\
	ScribbleSup\textdagger 	{\scriptsize\cite{scribblesup}}			& {\scriptsize\textit{L.}+\textit{S.}} 	& 63.1 & -	  \\
	AISI 					{\scriptsize\cite{salient-instance}}		& {\scriptsize\textit{L.}+\textit{I.}} 	& 63.6 & 64.5 \\
	\hline
	
	CCNN\textdagger 		{\scriptsize\cite{ccnn}} 		 		& {\scriptsize\textit{L.}} 				& 35.3 & 35.6 \\
	EM-Adapt\textdagger 	{\scriptsize\cite{em-adapt}}	 				& {\scriptsize\textit{L.}}  			& 38.2 & 39.6 \\
	STC\textdagger 			{\scriptsize\cite{stc}}			 		& {\scriptsize\textit{L.}} 				& 49.8 & 51.2 \\
	SEC\textdagger 			{\scriptsize\cite{sec}}					& {\scriptsize\textit{L.}}				& 50.7 & 51.7 \\
	AugFeed\textdagger 		{\scriptsize\cite{augfeed}}				& {\scriptsize\textit{L.}}			  	& 54.3 & 55.5 \\	
	AE-PSL\textdagger 		{\scriptsize\cite{erasing}}	 			& {\scriptsize\textit{L.}}				& 55.0 & 55.7 \\
	GuidedSeg\textdagger 	{\scriptsize\cite{guided-seg}}			& {\scriptsize\textit{L.}}				& 55.7 & 56.7 \\
	DCSP 					{\scriptsize\cite{dcsp}}			 		& {\scriptsize\textit{L.}}				& 60.8 & 61.9 \\	
	AFFNet\textdagger 		{\scriptsize\cite{affinity-net}} 		& {\scriptsize\textit{L.}}				& 58.4 & 60.5 \\
	MDC\textdagger 			{\scriptsize\cite{multi-dilation}} 		& {\scriptsize\textit{L.}}				& 60.4 & 60.8 \\
	MCOF 					{\scriptsize\cite{iterative-mine}}		& {\scriptsize\textit{L.}} 				& 60.3 & 61.2 \\	
	DSRG 					{\scriptsize\cite{seed-growing}}			& {\scriptsize\textit{L.}} 				& 61.4 & 63.2 \\	
	SeeNet 					{\scriptsize\cite{seenet}}		 		& {\scriptsize\textit{L.}} 				& 63.1 & 62.8 \\
	
	\hline
	\multicolumn{4}{l}{\textit{Ours:}} \\
	CIAN (res50)										 			& {\scriptsize\textit{L.}} 				& 62.4 & \textbf{63.8} \\
	CIAN (res101)													& {\scriptsize\textit{L.}} 				& \textbf{64.3} & \textbf{65.3} \\
	\hline
\end{tabular}}

\caption{Comparison of state-of-the-arts on VOC 2012. Methods marked by \textdagger\;use VGG16, the others use ResNet101. The supervision (Sup.) includes: image-level label (\textit{L.}), instance saliency (\textit{I.}), bounding box (\textit{B.}), scribble (\textit{S.}) and full supervision (\textit{F.}).}
\label{table:state-of-the-art}
\end{table}

\begin{figure}[t]
\begin{center}
	\includegraphics[width=0.9\linewidth]{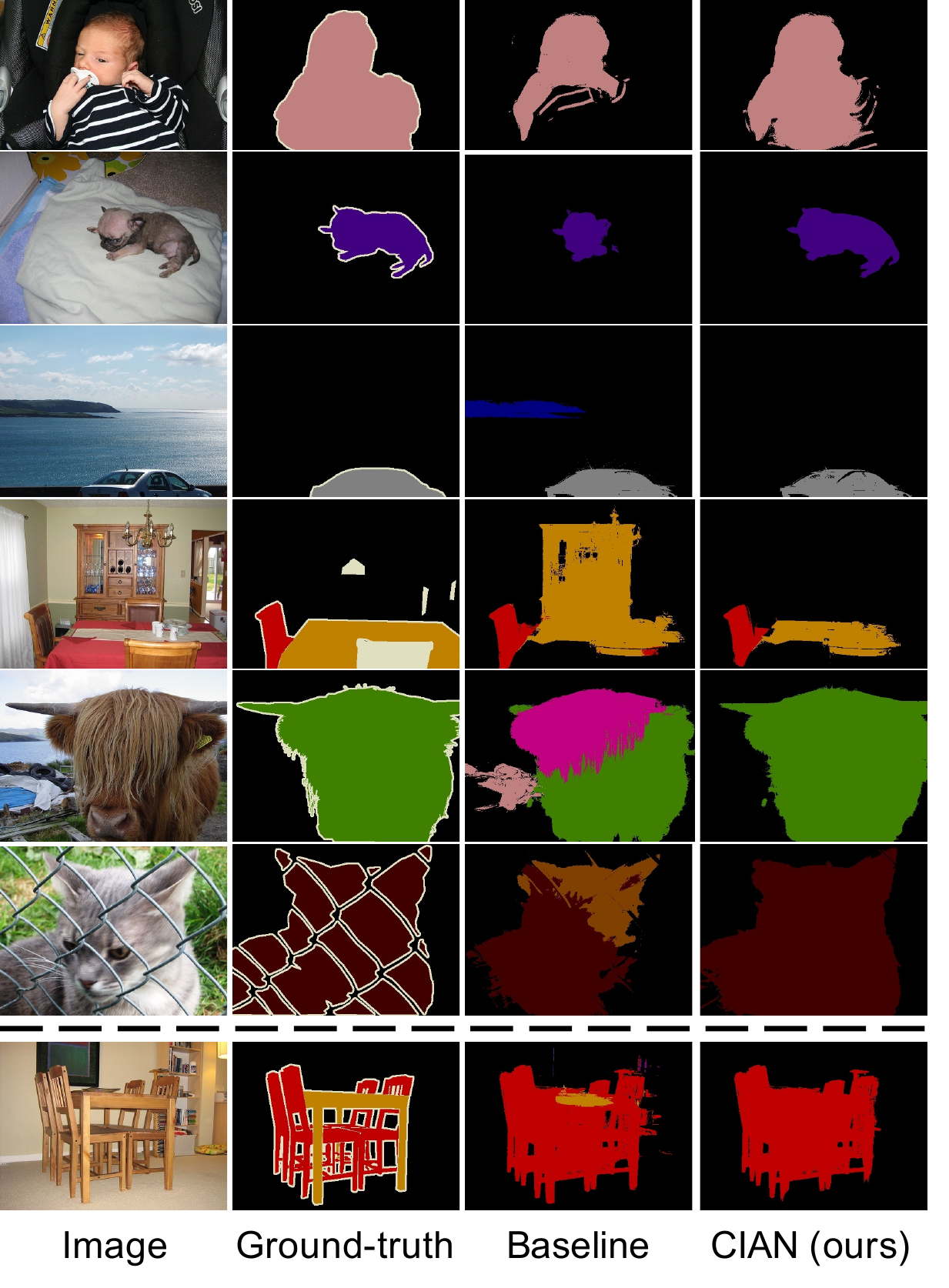}
\end{center}
\caption{Visualization of the predicted results on VOC 2012 val set. The first six rows show the typical cases to demonstrate how the CIAN outperforms the baseline. The last row shows a typical failure case where multiple classes are of similar appearance. Best viewed in color.}
\label{fig:predictions}
\end{figure}

\subsection{Qualitative Results}
To help better understand the final effect of the cross-image affinity module to the predictions, we visualize some of the typical predictions of both the baseline and our CIAN, as shown in Fig. \ref{fig:predictions}.
 The first two rows show that by the CIAN module that borrows information from other images, some of the missing parts can be completed. The next two rows show that some false positive predictions can be inhibited because they are never matched in any reference images. The following two rows demonstrate that the module helps to exclude clutter. This is because cross-image relationships help the network to learn more consistent representations across the whole dataset, and the related representations can be strengthened by each other, thus reduces the noise. The last row shows a typical failure case that interweaving objects with similar appearance and small spatial scale are confused and wrongly optimized. We leave it for future studies to address this problem.

% %%%%% SECTION5:conclusion

\section{Conclusion}
In this paper, we propose to leverage cross-image relationships for weakly supervised semantic segmentation. We propose an end-to-end CIAN module to build pixel-level affinities across different images, which can be directly plugged into existing segmentation networks. With the help of cross-image relationships, incomplete regions can retrieve supplementary information from other images to obtain more integral object region estimations and rectify the ambiguity. Extensive experiments demonstrate the advantage of utilizing cross-image relationships. Besides, our approach achieves state-of-the-art performance on VOC 2012 semantic segmentation task with only image-level labels.

% %%%%% SECTION6:Acknowledgements
\section{Acknowledgements}
This work was supported in part by the National Key R\&D Program of China (No. 2018YFB1402600, No. 2018YFB1402605), the National Natural Science Foundation of China (No. 61836014, No. 61761146004, No. 61773375, No. 61721004), the Beijing Municipal Natural Science Foundation (No. Z181100008918010) and CAS-AIR.

\bibliographystyle{aaai}
\bibliography{ref.bib}

\end{document}